\documentclass[sigconf]{acmart}

\AtBeginDocument{%
  \providecommand\BibTeX{{%
    \normalfont B\kern-0.5em{\scshape i\kern-0.25em b}\kern-0.8em\TeX}}}

\copyrightyear{2025}
\acmYear{2025}
\setcopyright{cc}
\setcctype{by-nc-sa}
\acmConference[MM '25]{Proceedings of the 33rd ACM International Conference on Multimedia}{October 27--31, 2025}{Dublin, Ireland}
\acmBooktitle{Proceedings of the 33rd ACM International Conference on Multimedia (MM '25), October 27--31, 2025, Dublin, Ireland}\acmDOI{10.1145/3746027.3758280}
\acmISBN{979-8-4007-2035-2/2025/10}

\usepackage{xcolor}
\usepackage{multirow}
\usepackage{subcaption}

\begin{document}

\title{MFFI: Multi-Dimensional Face Forgery Image Dataset for Real-World Scenarios}


\author{Changtao Miao}
\authornote{These authors contributed equally to this paper.}
\author{Yi Zhang}
\authornotemark[1]
\author{Man Luo}
\authornotemark[1]
\affiliation{%
  \institution{Ant Group}
  \city{Hangzhou}
  \country{China}
}

\author{Weiwei Feng}
\author{Kaiyuan Zheng}
\affiliation{%
  \institution{Ant Group}
  \city{Hangzhou}
  \country{China}
}

\author{Qi Chu}
\author{Tao Gong}
\affiliation{%
  \institution{Anhui Province Key Laboratory of Digital Security}
  \city{Hefei}
  \country{China}}

\author{Jianshu Li}
\authornote{Project leader. Email:jianshu.l@antgroup.com}
\affiliation{%
  \institution{Ant Group}
  \city{Hangzhou}
  \country{China}
}

\author{Yunfeng Diao}
\authornote{Corresponding authors. Email:diaoyunfeng@hfut.edu.cn; xshao@baai.ac.cn}
\affiliation{%
  \institution{Hefei University of Technology}
  \city{Hefei}
  \country{China}
}

\author{Wei Zhou}
\affiliation{%
  \institution{Cardiff University}
  \city{Cardiff}
  \country{UK}}

\author{Joey Tianyi Zhou}
\affiliation{%
  \institution{IHPC, Agency for Science, Technology and Research}
  \institution{CFAR, Agency for Science, Technology and Research}
  \country{Singapore}}

\author{Xiaoshuai Hao}
\authornotemark[3]
\affiliation{%
 \institution{Beijing Academy of Artificial Intelligence}
 \city{Beijing}
 \country{China}}

\renewcommand{\shortauthors}{Changtao Miao et al.}

\begin{abstract} 
Rapid advances in Artificial Intelligence Generated Content (AIGC) have enabled increasingly sophisticated face forgeries, posing a significant threat to social security. 
However, current Deepfake detection methods are limited by constraints in existing datasets, which lack the diversity necessary in real-world scenarios. 
Specifically, these data sets fall short in four key areas: unknown of advanced forgery techniques, variability of facial scenes, richness of real data, and degradation of real-world propagation. 
To address these challenges, we propose the Multi-dimensional Face Forgery Image (\textbf{MFFI}) dataset, tailored for real-world scenarios. 
MFFI enhances realism based on four strategic dimensions: 1) Wider Forgery Methods; 2) Varied Facial Scenes; 3) Diversified Authentic Data; 4) Multi-level Degradation Operations. 
MFFI integrates $50$ different forgery methods and contains $1024K$ image samples. 
Benchmark evaluations show that MFFI outperforms existing public datasets in terms of scene complexity, cross-domain generalization capability, and detection difficulty gradients. 
These results validate the technical advance and practical utility of MFFI in simulating real-world conditions. 
The dataset and additional details are publicly available at {https://github.com/inclusionConf/MFFI}.
\end{abstract}

\ccsdesc[500]{Computing methodologies~Computer vision}
\ccsdesc[300]{Computing methodologies~Biometrics}

\keywords{Deepfake, Face Forgery Detection, Real-World Scenarios}



\maketitle

\section{Introduction}

\begin{figure}[!t]
\centering
\includegraphics[width=0.48\textwidth]{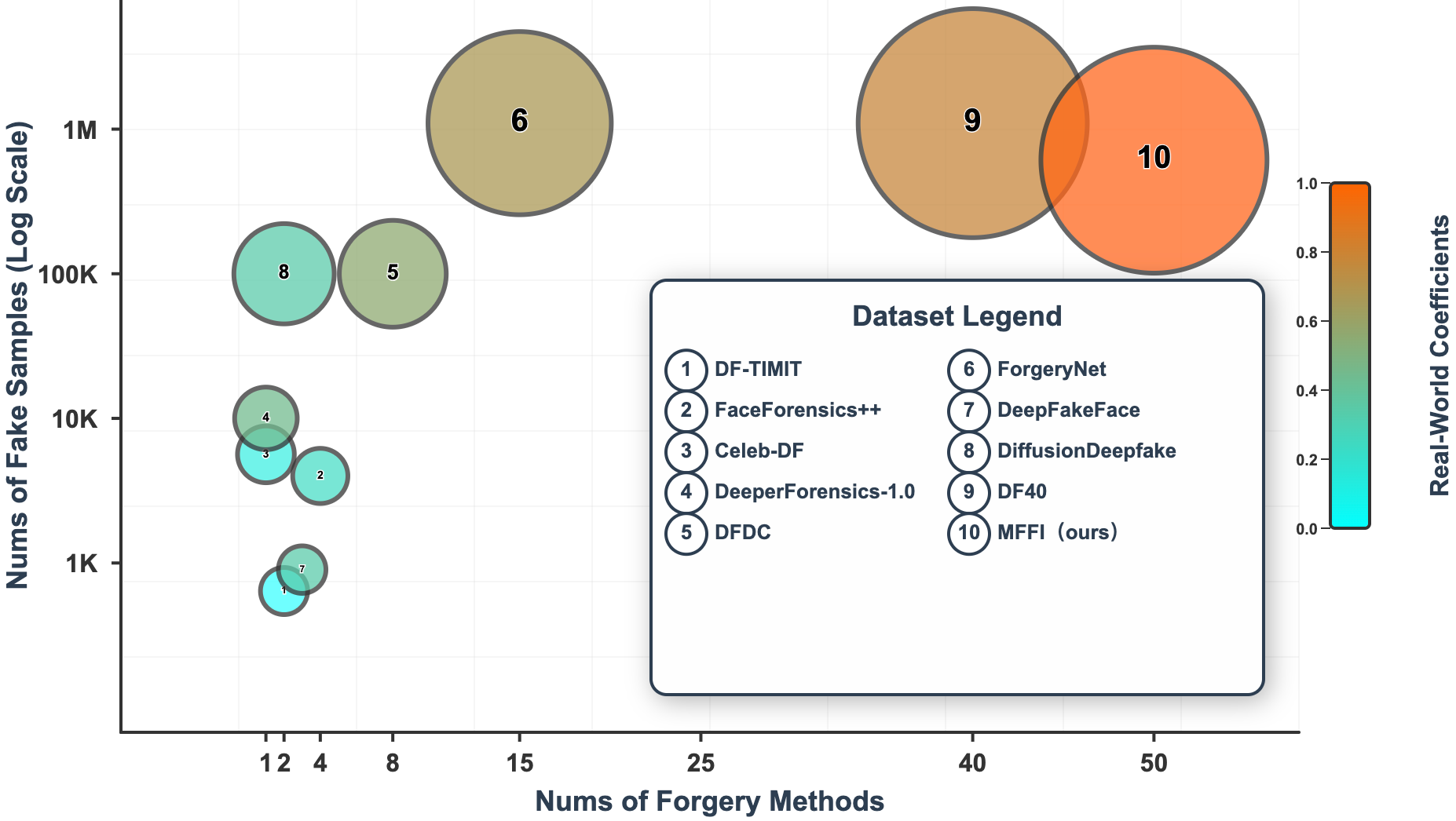}
\caption{
Comparison with other dataset in diversity and real-world realism.
The \textit{Real-World Coefficients} denote the degree of similarity between datasets and real-world scenarios.
}
\label{fig:intro_1}
\vspace{-0.2cm}
\end{figure}

\begin{figure*}[!t]
\centering
\includegraphics[width=0.98\textwidth]{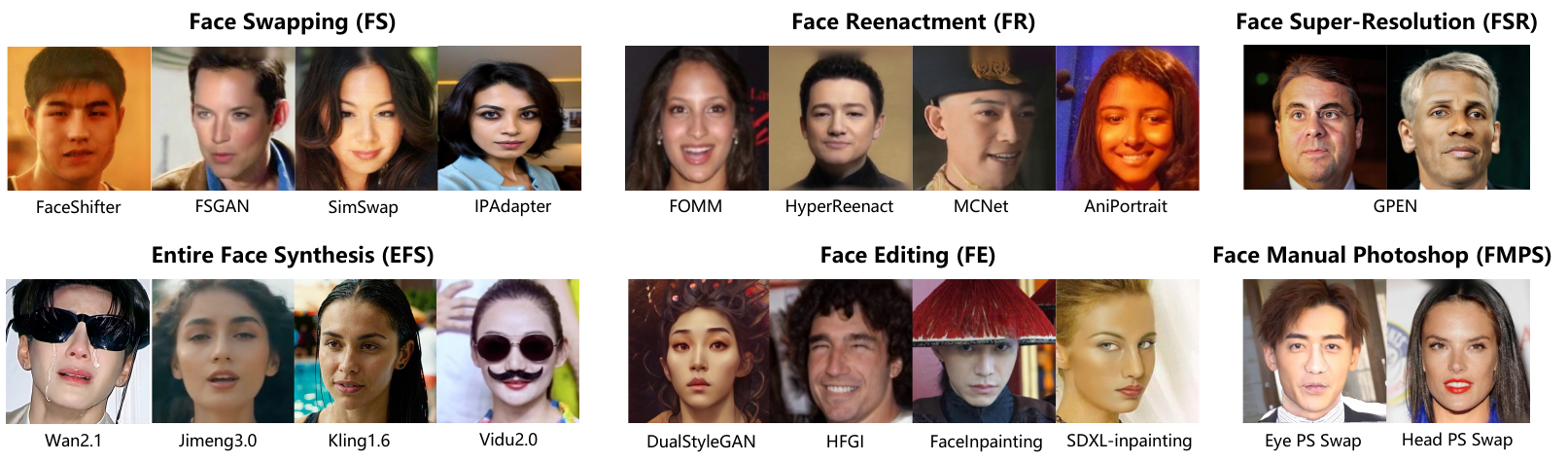}
\caption{
Examples of generated images from 6 major face forgery categories in our MFFI dataset. The categories include Face Swapping (FS, 15 types), Face Reenactment (FR, 8 types), Entire Face Synthesis (EFS, 13 types), Face Editing (FE, 6 types), Face Super-Resolution (FSR, 1 type), and Face Manual Photoshop (FMPS, 7 types).
}
\label{fig:dataset_2}
\end{figure*}

As we accelerate towards the era of Artificial General Intelligence, its core technology, Artificial Intelligence Generated Content (AIGC), demonstrates remarkable capabilities in producing hyper-realistic content, such as highly lifelike text, images, and audio-visual content~\cite{cao2025survey}.
While AIGC expands creative possibilities, it simultaneously presents unprecedented challenges to societal governance structures. Central among these threats are deepfake-based face forgery attacks~\cite{li2020celeb,rossler2019faceforensics++,yandf40}, where identity features can be precisely replicated, facial microexpressions altered, or social trust mechanisms systematically eroded. 
Such capabilities provide technical enablers for malicious activities such as financial fraud and political subversion. 
In response, face forgery detection technologies~\cite{miao2021learning,zhuang2022towards,miao2022hierarchical,miao2023f,zhuang2022uia,tan2022transformer,miao2023multi} have emerged as a critical research domain to safeguard the integrity of digital societal infrastructures. 

However, current deepfake detection methods~\cite{kong2022detect,luo2023beyond} exhibit inadequate generalization ability to real-world scenarios due to limitations in training data quality and diversity. 
This shortfall arises primarily from the multi-dimensional complexity of forged samples in authentic environments: 
(1) \textbf{Unknown of Advanced Forgery Techniques}: attackers frequently use emerging forgery techniques that are unseen during training~\cite{nadimpalli2022improving};
(2) \textbf{Variability of Facial Scenes}: facial images often appear under varying environmental conditions and composition styles, which hinder the extraction of invariant features~\cite{zhang2024genface}; 
(3) \textbf{Richness of Real Data}: fake faces in real-world may originate from highly divergent source distribution~\cite{dolhansky2020deepfake}, leading to domain generalization gaps for models trained on narrow datasets; 
(4) \textbf{Degradation of Real-world Propagation}: transmission over social platforms often degrade input quality via adversarial perturbations, color distortions, and compression-induced blurring~\cite{zhang2025adversarially,diao2024vulnerabilities,zhu2023genimage,diao2025tasar}. 
While existing Deepfake datasets address subsets of these challenges, a unified benchmark integrating all four core dimensions remains unexplored. For example, FF++~\cite{rossler2019faceforensics++} and Celeb-DF~\cite{li2020celeb} are constrained by the singularity of generation techniques and the limitations of the original data.
DFDC~\cite{dolhansky2020deepfake} relies on 8 traditional manipulation methods, lacking coverage of advanced techniques like diffusion models~\cite{ho2020denoising}. The latest DF40~\cite{narayan2023df} increases the generation diversity to 40 methods but reuses real videos from FF++ and Celeb-DF, limiting the expansion of the source domain. 
Moreover, these datasets largely overlook real-world transmission artifacts such as color shifts, compression, blurring, and adversarial perturbations. Therefore, developing a comprehensive face forgery dataset that integrates all four core dimensions is both urgent and essential to effectively address real-world challenges.

\begin{table*}[tb]
\begin{center}

\caption{Comparison of previous face forgery datasets. MFFI surpasses any other dataset in Wider Forgery Methods (WFM), Varied Facial Scenes (VFS), Diversified Authentic Data (DAD), and Multi-level Degradation Operations (MDO). 
The \textit{Latest Face Foregry} represents the latest forgery method in this dataset.
The \textit{Fake Samples} represents the number of forged samples. Each forged video or image is considered as one sample.
}
\label{tab:datasets_1}
\resizebox{0.98\linewidth}{!}{
\begin{tabular}{l|ccccccccccccc}
\toprule[1pt]
\multirow{2}{*}{Dataset} &
  \multirow{2}{*}{Publication} &
  \multirow{2}{*}{\begin{tabular}[c]{@{}c@{}}Latest \\ Face Forgery\end{tabular}} &
  \multicolumn{8}{c}{Wider Forgery Methods (WFM)} &
  \multirow{2}{*}{VFS} &
  \multirow{2}{*}{DAD} &
  \multirow{2}{*}{MDO} \\ \cline{4-11}
                    &            &                         & Methods & FS & FR & EFS & FE & FSR & FMPS & Fake Smaples &     &     &     \\ \hline
DF-TIMIT~\cite{korshunov2018deepfakes}            & ArXiv'18   & faceswap-GAN~\cite{faceswapgan2019} (2018)     & 2       & 2  & -  & -   & -  & -   & -    & 640          & -   & -   & -   \\
FaceForensics++~\cite{rossler2019faceforensics++}     & ICCV'19    & NeuralTextures (2019)   & 4       & 2  & 2  & -   & -  & -   & -    & 4,000        & -   & -   & -   \\
Celeb-DF~\cite{li2020celeb}            & CVPR'20    & Unknown                 & 1       & 1  & -  & -   & -  & -   & -    & 5,639        & -   & -   & -   \\
DeeperForensics-1.0~\cite{Deeperforensics} & CVPR'20    & DF-VAE~\cite{Deeperforensics} (2020)           & 1       & 1  & -  & -   & -  & -   & -    & 10K          & -   & -   & \checkmark \\
DFDC~\cite{dolhansky2020deepfake}                & ArXiv'20   & StyleGAN~\cite{karras2019style} (2018)         & 8       & 5  & 1  & 2   & -  & -   & -    & 0.1M+        & -   & \checkmark & -   \\
ForgeryNet~\cite{he2021forgerynet}          & CVPR'21    & StarGANv2~\cite{karras2020analyzing} (2020)        & 15      & 6  & 4  & 2   & 3  & -   & -    & 1.1M+        & -   & -   & -   \\
FakeAVCeleb~\cite{FakeAVCeleb}         & NeurIPS'21 & Wav2Lip~\cite{prajwal2020lip} (2021)          & 4       & 2  & 2  & -   & -  & -   & -    & 9,500        & -   & -   & -   \\
DF3~\cite{DF3}                 & TMM'22     & StyleGAN3~\cite{karras2021alias} (2021)        & 6       & -  & -  & 6   & -  & -   & -    & 15k+         & -   & -   & -   \\
DeepFakeFace~\cite{song2023robustness}        & ArXiv'23   & Stable-Diffusion~\cite{sd} (2021) & 3       & 1  & -  & 2   & -  & -   & -    & 90K          & -   & -   & -   \\
DiffusionDeepfake~\cite{bhattacharyya2024diffusion}   & ArXiv'24   & Stable-Diffusion~\cite{sd} (2021) & 2       & -  & -  & 2   & -  & -   & -    & 0.1M+        & -   & -   & -   \\
DF40~\cite{yandf40}                & NeurIPS' 24   & PixArt-$\alpha$~\cite{chenpixart} (2024)         & 40      & 10 & 13 & 12  & 5  & -   & -    & 1.1M+        & -   & -   & -   \\ \hline
\textbf{MFFI (ours)} & -          & \textbf{Jimeng3.0 \cite{Jimeng2024Video} (2025)}       & \textbf{50}      & 15 & 8  & 13  & 6  & 1   & 7    & 0.7M+        & \checkmark & \checkmark & \checkmark \\ 
\bottomrule[1pt]
\end{tabular}
}
\end{center}
\end{table*}

To address these challenges, we introduce the Multi-dimensional Face Forgery Image (\textbf{MFFI}) dataset, which designed to enhance the adaptability of face forgery detection models in real-world scenarios. MFFI systematically covers all the four complex forgery dimensions that may exist in realistic settings, with core innovations in the following dimensions: 

\noindent \textbf{Wider Forgery Methods:} Beyond traditional Deepfake techniques, we innovatively incorporate image super-resolution and manual Photoshop. In particular, we integrate commercial tools just released in 2025 (e.g., Wan2.1~\cite{wan}, Vidu2.0~\cite{vidu}, Kling1.6~\cite{KlingAI}, and Jimeng2.0~\cite{Jimeng2024Video}) to ensure comprehensive coverage of the latest techniques. Overall, MFFI incorporates up to 50 diverse forgery methods, surpassing current datasets by a big margin.

\noindent \textbf{Varied Facial Scenes:} This dataset comprehensively covers different skin tones, age groups, pose angles, and real shooting scenarios including occlusions, complex lighting, and diverse backgrounds, fully simulating the complexity of facial images in the real world.

\noindent \textbf{Diversified Authentic Data:} We integrate real facial data from four distinct sources, breaking away from the single-source collection mode of existing datasets to ensure diversity and authenticity in the foundational data.

\noindent \textbf{Multi-level Degradation Operations:} To simulate potential degradation processes during data propagation, we randomly embed multi-level interference factors including blur processing, noise interference, color distortion, geometric deformation, sharpening artifacts, compression distortion, and adversarial perturbations.

A detailed comparison with current face forgery datasets are shown in Table \ref{tab:datasets_1}. To our best knowledge, MFFI is the first face forgery dataset to uniformly consider all the four strategic dimensions (WFM, VFS, DAD, and MDO) in the real-world scenarios. Through this multi-dimensional data construction strategy, we achieve a dataset encompassing $50$ diverse forgery techniques and $1024K$ samples (including both real and forged faces) to simulate forged faces in real-world environments. To validate the practical value of our dataset, we conduct extensive generalization and robustness experiments to analyze the performance of existing detection methods.

By addressing the critical challenge face forgery in real-world scenarios, our MFFI dataset aims to drive advancements in the development of robust, real-world-oriented face forgery detection methodologies.
To do this end, the MFFI dataset served as the core benchmark for the Global Multimedia Deepfake Detection Challenge \footnotemark[1] held in Kaggle~\cite{zhang2024inclusion}, attracting 1500 teams globally. 
\footnotetext[1]{https://www.kaggle.com/competitions/multi-ffdi}

\section{Related Work}
\subsection{Face Forgery Datasets}
Early datasets ~\cite{rossler2019faceforensics++,li2020celeb} were constrained by scene singularity, using single-source domain data and limited face manipulation techniques. While subsequent datasets like DFDC~\cite{dolhansky2020deepfake} enhanced data diversity, they still used only a few generation techniques, failing to overcome technological coverage limitations.
ForgeryNet~\cite{he2021forgerynet} achieved large-scale dataset construction but neglected recent diffusion model-based generation paradigms. DeepFakeFace~\cite{song2023robustness} and DiffusionDeepfake~\cite{bhattacharyya2024diffusion} datasets introduced cutting-edge diffusion model technologies but suffered from homogenization of forgery modes. DF40~\cite{yandf40} improved diversity by expanding generation techniques, but was limited by existing original data (i.e., FF++~\cite{rossler2019faceforensics++} and Celeb-DF~\cite{li2020celeb}) distribution constraints and overlooked real-world interference and degradation factors.
To address these deficiencies, our MFFI dataset effectively resolves the aforementioned limitations by simulating real-world scenarios across four dimensions.

\subsection{Face Forgery Methods}
Existing face forgery research typically categorizes forgery methods into four types: Face Swapping (FS), Face Reenactment (FR), Entire Face Synthesis (EFS), and Face Editing (FE). 
FS and FR are the most common technologies \cite{deepfake2019,faceswap2019,thies2016face2face,thies2019deferred} in face forgery datasets. However, with the rapid development of generative technologies, these techniques often suffer from obsolescence and homogenization. 
To address this issue, our MFFI dataset incorporates 15 different FS methods and 8 diverse FR methods, including latest audio-driven technologies, such as AniPortrait~\cite{wei2024aniportrait}, InstantID~\cite{wang2024instantid}.
EFS techniques have matured significantly due to progress in diffusion models~\cite{bhattacharyya2024diffusion}. 
The MFFI incorporates the latest commercial text-to-video and image-to-video models to ensure technological frontier coverage, e.g., Wan2.1~\cite{wan} and Jimeng3.0~\cite{Jimeng2024Video}. 
FE represents a more fine-grained form of forgery. The MFFI implements both traditional GAN-based \cite{goodfellow2014generative} and the latest diffusion~\cite{ho2020denoising} model-based methods in this category.
While these classifications encompass most mainstream face forgery methods, they overlook two common forms of facial manipulation: Face Super-Resolution (FSR) and Face Manual PhotoShop (FMPS). 
Therefore, MFFI supplements 1 FSR technique and 7 FMPS operations to enhance the dataset's comprehensiveness.

\section{Dataset}

\subsection{Overview} \label{sec_overview}
The MFFI dataset addresses existing limitations through four dimensions, i.e., WFM, VFS, DAD, and MDO. 
First, WFM expands beyond conventional categories (FS, FR, EFS, FE) to include Face Super-Resolution (FSR) and Face Manual Photoshop (FMPS) operations. 
Second, VFS systematically incorporates skin tones, head poses, occlusions, backgrounds, age ranges, and illumination conditions to simulate real-world complexity. 
Third, DAD combines established benchmarks with newly collected celebrity face images. 
Fourth, MDO simulates internet transmission artifacts (compression, blurring) while incorporating adversarial attacks for environmental fidelity.
To prioritize real-world complexity, MFFI focuses on image-level data provision. For video-based forgery methods, forged frames are analyzed through frame sampling. 
As shown in Table \ref{tab:datasets_2}, MFFI includes Train, Valid, Test, and Test-D sets.

\begin{figure}[!t]
\centering
\includegraphics[width=0.47\textwidth]{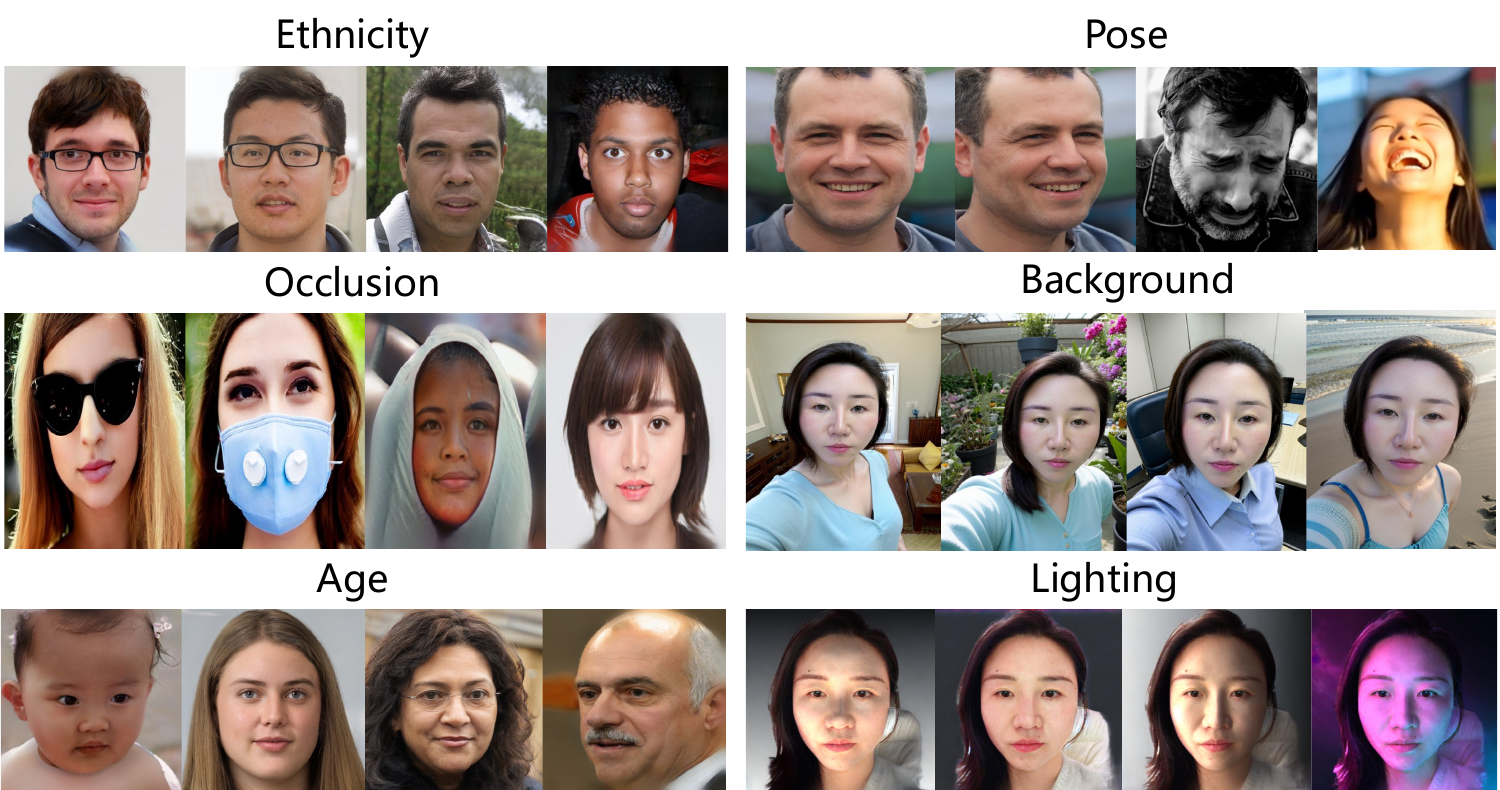}
\caption{
Fake Examples of the VFS dimension. 
}
\label{fig:dataset_3}
\vspace{-0.2cm}
\end{figure}

\subsection{Wider Forgery Methods}
We introduce and implement 50 diverse face forgery techniques to ensure the realization of the diversity objective. These varied forgery techniques are categorized into six main classes:

\textbf{Face Swapping (FS):} This technique replaces the target face with the source face while preserving the identity of the source and the background of the target. 
We implement 15 different methods, including 
SimSwap~\cite{chen2020simswap}, 
FaceShifter~\cite{li2019faceshifter}, 
FaceFusion \cite{facefusion2023},
FSGAN~\cite{nirkin2019fsgan}, 
InfoSwap~\cite{gao2021information}, 
Stable-Diffusion-1.5 \cite{sd}, 
HiFiface~\cite{wang2021hififace}, 
IPAdapter~\cite{ye2023ip}, 
MegaFS~\cite{zhu2021one}, 
MobileFaceSwap~\cite{xu2022mobilefaceswap}, 
FaceMerge~\cite{FaceMerge2}, 
RAFSwap~\cite{xu2022region}, 
e4s~\cite{li2023e4s}, AIM \cite{li2022multi}, I2G \cite{zhao2021learning}, and SBI \cite{shiohara2022detecting}.

\textbf{Face Reenactment (FR):} It modifies the motion or expression information of the original face using video, audio, or text-driven approaches. 
We implement 8 methods, including FOMM~\cite{zhang2021facial}, MCNet~\cite{hong2023implicit}, ArticulatedAnimation~\cite{siarohin2021motion}, DaGAN~\cite{hong2022depth}, HyperReenact~\cite{bounareli2023hyperreenact}, MonkeyNet~\cite{bala2023monkeynet}, AniPortrait~\cite{wei2024aniportrait}, and Wav2Lip~\cite{prajwal2020lip}.

\textbf{Entire Face Synthesis (EFS):} This category uses GANs or diffusion models to directly generate complete synthetic faces. 
We implement 13 techniques, including StyleGAN2 \cite{karras2019style}, StyleGAN3 \cite{zhu2023stylegan3},
InstantID~\cite{wang2024instantid}, Stable-Diffusion-XL \cite{podellsdxl}, Hailuoai~\cite{hailuoai}, Jimeng2.0~\cite{Jimeng2024Video}, Jimeng3.0~\cite{Jimeng2024Video}, Kling1.6~\cite{KlingAI}, Pika1.5~\cite{MellisIncPika2024v1.5}, Pixverse~\cite{PixVerseAI2025}, Vidu2.0~\cite{vidu}, Vivago~\cite{VivagoAI2025}, and Wan2.1~\cite{wan}.

\textbf{Face Editing (FE):} This class aims to modify specific attributes of a given facial image. 
We implement 6 techniques, including HFGI~\cite{wang2022high}, pSp~\cite{richardson2021encoding}, FaceInpainting \cite{luo2023reference}, DualStyleGAN~\cite{yang2022pastiche}, SD-Inpainting~\cite{sd}, and SD-XL-Inpainting~\cite{podellsdxl}.

\textbf{Face Super-Resolution (FSR):} It enhances image quality by recovering details from low-resolution fake face images. 
We implemented the commonly used GPEN~\cite{yang2021gan} technique for this.

\textbf{Face Manual Photoshop (FMPS):} This category involves manual manipulation of local facial areas using Photoshop software. We implement 6 common operations, including 
Random Facial Component Swapping, 
Random Facial Component Excision, 
Full Face Swapping,
Facial Artifact Injection,
Random Region Erasure,
and Background Replacement.
We also apply 1 Image-Filtering operation (including Kuwahara, Bilateral, and Oil filters) to process forged faces and render them with artistic stylization, thereby simulating Photoshop-style manipulations of real-world scenarios.

All these forgery techniques were implemented based on corresponding research papers or common technical approaches, generating a rich set of forged facial images that comprehensively simulate diverse scenarios in real-world settings.

\begin{table}[tb]
\begin{center}
\caption{Composition of real/fake class and forgery types in the proposed MFFI dataset.
The \textit{Test-D} denotes the test set processed according to MDO.
}
\label{tab:datasets_2}
\resizebox{0.99\linewidth}{!}{
\begin{tabular}{cc|cccc|c}
\toprule[1pt]
\multicolumn{2}{c|}{Sample Type}                     & Train & Valid & Test  & Test-D & Total \\ \hline
\multirow{4}{*}{Real} & CelebA~\cite{liu2018large}                      & 100K  & -     & -     & -      & 100K  \\
                      & RFW~\cite{wang2019racial,wang2020mitigating}                          & -     & 60K   & 25K   & 25K    & 110K  \\
                      & CASIA-WebFace~\cite{yi2014learning}                & -     & -     & 25K   & 25K    & 50K   \\
                      & Chinese Celeb                & -     & -     & 30K   & 30K    & 60K   \\ \hline
\multirow{6}{*}{Fake} & FS          & 96K   & 53K   & 21K   & 21K    & 191K  \\
                      & FR        & 90K   & 2K    & 18K   & 18K    & 128K  \\
                      & EFS  & 9K    & -     & 14K   & 14K    & 37K   \\
                      & FE           & 16K   & 2K    & 15.2K & 15.2K  & 48.4K \\
                      & FSR  & 198K  & 23K   & 4.8K  & 4.8K   & 230.6 \\
                      & FMPS & 15K   & -     & 27K   & 27K    & 69K   \\ \hline
\multicolumn{2}{c|}{Total}                           & 524K  & 140K  & 180K  & 180K   & \textbf{1024K} \\ 
\bottomrule[1pt]
\end{tabular}
}
\end{center}
\vspace{-0.2cm}
\end{table}

\subsection{Varied Facial Scenes}
To further enhance the facial scenes, we filtered the aforementioned authentic and forged face images across six dimensions, specifically:
1) \textbf{Ethnicity:} Covering Caucasian, African, Indian, and Asian populations.
2) \textbf{Age:} Spanning from 0 to 85 years old.
3) \textbf{Pose:} Including frontal, profile, downward, and upward views.
4) \textbf{Occlusion:} Considering scenarios with sunglasses, face masks, hoodies, and bangs.
5) \textbf{Background:} Distinguishing between indoor and outdoor scenes.
6) \textbf{Lighting:} Incorporating conditions of strong light, low light, and colored lighting.

As illustrated in Figure \ref{fig:dataset_3}, this comprehensive approach to facial scene design captures the diversity and complexity of real-world facial data, providing robust support for the diversity and realism of the dataset.

\subsection{Diversified Authentic Data}
In face forgery datasets, the diversity of authentic facial data is crucial. However, previous datasets often lacked in this aspect, potentially leading to biases in model training. To construct a dataset better aligned with real-world scenarios, 
we select three distinctive real-world face recognition and a self-collected celebrity datasets:

\textbf{CelebA~\cite{liu2018large}:} A collection of celebrity images from the internet, featuring $200K$ images across $10K$ identities, providing a rich sample size and diverse identities.

\textbf{Racial Faces in the Wild (RFW)~\cite{wang2019racial}:} Meticulously curated to represent four racial groups - Caucasian, Indian, Asian, and African. Each racial subset comprises $10K$ images spanning $3K$ identities, ensuring balanced ethnic representation.
To further mitigate racial bias, we incorporate the extended variants of the RFW dataset: the BUPT-GlobalFace and BUPT-BalancedFace datasets \cite{wang2020mitigating}.

\textbf{CASIA-WebFace~\cite{yi2014learning}:} A large-scale face dataset with about $500K$ facial images across $10K$ identities, offering enhanced sample quantity and identity coverage.

\textbf{Chinese Celeb:} To address the under-representation of Chinese internet facial data, we self-collect Chinese celebrity faces containing $30K$ images and about $1.5K$ identities, further enhancing geographical diversity.

\begin{table*}[tb]
\begin{center}
\caption{Intra-dataset evaluation with the SOTAs on MFFI datasets. 
}
\label{tab:intra-datasets_1}
\resizebox{0.80\linewidth}{!}
{
\begin{tabular}{c|c|cccc|cccc}
\toprule[1pt]
\multirow{2}{*}{Types} & \multirow{2}{*}{Models} & \multicolumn{4}{c|}{Test}         & \multicolumn{4}{c}{Test-D}        \\ \cline{3-10} 
                       &                         & ACC    & AUC    & EER    & AP     & ACC    & AUC    & EER    & AP     \\ \hline
\multirow{2}{*}{\begin{tabular}[c]{@{}c@{}}Spatial \\ Detector\end{tabular}} 
& Xception \cite{chollet2017xception} &
  0.7683 & 0.8522 & 0.2320 & 0.8855 & 0.6470 & 0.7075 & 0.3550 & 0.7727 \\
& RFM \cite{wang2021representative} & 
    0.7588 & 0.8487 & 0.2336 & 0.8853 & 0.6560 & 0.7203 & 0.3441 & 0.7869 \\ \hline
\multirow{2}{*}{\begin{tabular}[c]{@{}c@{}}Frequency \\ Detector\end{tabular}} &
  SRM \cite{luo2021generalizing} & 0.7619 & 0.8765 & 0.2105 & 0.9146 & 0.5524 & 0.5801 & 0.4307 & 0.6590 \\
    & SPSL \cite{liu2021spatial} & 0.7492 & 0.8455 & 0.2396 & 0.8772 & 0.6305 & 0.7144 & 0.3486 & 0.7763 \\ 
\bottomrule[1pt]
\end{tabular}
}
\end{center}
\end{table*}

\subsection{Multi-level Degradation Operations}
To simulate the degradation of the image during transmission, we randomly applied various disturbance operations, primarily categorized into conventional disturbances and deep adversarial attacks. 

For conventional disturbances, we employ common perturbation techniques including uniform blur, Gaussian blur, uniform noise, Gaussian noise, sharpening, compression, and geometric transformations. These operations are widely observed in data transformation processes across social networks on the Internet, effectively mimicking real-world scenarios. 

In terms of deep adversarial attacks, to further enhance the realism of forgery scenarios and counter potential malicious attackers, we simulate adversarial attack strategies. Specifically, we generate perturbations using PatchAttack~\cite{yang2020patchattack}, a black-box attack method based on reinforcement learning. This approach induces model misclassifications by superimposing small textured patches onto images, thereby increasing the difficulty of face forgery detection.

To align with real-world evaluation requirements, we apply these degradation operations exclusively to the test set (i.e., Test-D). By incorporating both conventional disturbances and advanced adversarial attacks, our methodology provides a comprehensive framework for assessing the robustness of face forgery detection models under diverse and challenging conditions.

\begin{table*}[tb]
\begin{center}
\caption{Cross-datasets evaluation with the SOTAs. The models train on DF40-FS \cite{yandf40} or FF++(C23) \cite{rossler2019faceforensics++}, then they test on Test and Test-D sets of ours MFFI datasets.
}
\label{tab:cross-datasets_1}
\resizebox{0.8\linewidth}{!}
{
\begin{tabular}{c|c|cccc|cccc}
\toprule[1pt]
\multirow{2}{*}{Models} & \multirow{2}{*}{Train Set}  & \multicolumn{4}{c|}{Test}         & \multicolumn{4}{c}{Test-D}        \\ \cline{3-10} 
 &
   &
  \multicolumn{1}{c}{ACC} &
  \multicolumn{1}{c}{AUC} &
  \multicolumn{1}{c}{EER} &
  \multicolumn{1}{c|}{AP} &
  \multicolumn{1}{c}{ACC} &
  \multicolumn{1}{c}{AUC} &
  \multicolumn{1}{c}{EER} &
  \multicolumn{1}{c}{AP} \\ \hline
Xception \cite{chollet2017xception} & \multirow{4}{*}{DF40-FS \cite{yandf40}}               & 0.4840 & 0.4571 & 0.5309 & 0.5588 & 0.4987 & 0.4741 & 0.5136 & 0.5648 \\
RFM \cite{wang2021representative} &  & 0.5285 & 0.4772 & 0.5219 & 0.5763 & 0.5332 & 0.4718 & 0.5226 & 0.5678 \\
SRM \cite{luo2021generalizing} &  & 0.5067 & 0.4540 & 0.5368 & 0.5591 & 0.5134 & 0.4697 & 0.5236 & 0.5620 \\
SPSL \cite{liu2021spatial} & & 0.4919 & 0.4584 & 0.5323 & 0.5707 & 0.5106 & 0.4795 & 0.5134 & 0.5748 \\ \hline
Xception \cite{chollet2017xception} & \multirow{4}{*}{FF++ (C23) \cite{rossler2019faceforensics++}}   & 0.5527 & 0.6249 & 0.4059 & 0.6905 & 0.5118 & 0.5118 & 0.4591 & 0.6418 \\
RFM \cite{wang2021representative} &  & 0.5282 & 0.6175 & 0.4161 & 0.6837 & 0.5169 & 0.5804 & 0.4412 & 0.6463 \\
SRM \cite{luo2021generalizing} &  & 0.6041 & 0.6552 & 0.3930 & 0.7148 & 0.5662 & 0.5914 & 0.4370 & 0.6564 \\
SPSL \cite{liu2021spatial}                    &                             & 0.5536 & 0.5564 & 0.4619 & 0.6416 & 0.5356 & 0.5429 & 0.4737 & 0.6227 \\ 
\bottomrule[1pt]
\end{tabular}
}
\end{center}
\end{table*}

\begin{table}[tb]
\begin{center}
\caption{Cross-datasets evaluation with the SOTAs. The models are trained on our MFFI, then they are tested on other datasets. The metric is AUC.
}
\label{tab:cross-datasets_2}
\resizebox{0.96\linewidth}{!}
{
\begin{tabular}{c|ccccc}
\toprule[1pt]
Moedels  & CDF-V1~\cite{li2020celeb} & CDF-V2~\cite{li2020celeb} & DFD~\cite{rossler2019faceforensics++}    & DFDC~\cite{dolhansky2020deepfake}   & Avg.   \\ \hline
Xception \cite{chollet2017xception} & 0.6933 & 0.6695 & 0.7117 & 0.5523 & 0.6567 \\
RFM \cite{wang2021representative}      & 0.6452 & 0.6346 & 0.6872 & 0.5375 & 0.6261 \\ \hline
SRM \cite{luo2021generalizing}     & 0.6932 & 0.5599 & 0.5921 & 0.5259 & 0.5928 \\
SPSL \cite{liu2021spatial}     & 0.7006 & 0.7030 & 0.7535 & 0.5942 & 0.6878 \\ 
\bottomrule[1pt]
\end{tabular}
}
\end{center}
\end{table}

\section{Evaluations}

\subsection{Experimental Setup}
In this study, all preprocessing and training codebases strictly adhere to the experimental setup of DeepfakeBench \cite{DeepfakeBench_YAN_NEURIPS2023}. We select the classic Xception \cite{chollet2017xception} as our baseline model and incorporate three SOTA detectors: SRM \cite{luo2021generalizing}, SPSL \cite{liu2021spatial}, and RFM \cite{wang2021representative}, all of which utilize Xception as their backbone network. 
This configuration enables a comprehensive analysis of traditional detection models' performance on our proposed MFFI dataset, facilitating a comparison of detection efficacy between MFFI and other datasets.
The algorithmic implementation details and training hyperparameters are consistent with the publicly available settings in DeepfakeBench \cite{DeepfakeBench_YAN_NEURIPS2023}, ensuring reproducibility and consistency of results. 

\subsection{Evaluation Protocols and Metrics}
To comprehensively evaluate the face forgery detection methods on the proposed MFFI dataset, we establish three evaluation protocols:
1) Intra-dataset; 2) Cross-dataset; 3) Zero-shot detection of MLLM.
Following the DeepfakeBench \cite{DeepfakeBench_YAN_NEURIPS2023}, we report ACC, AUC, EER, AP as metrics of the detection models.

\begin{figure*}[!t]
\centering
\includegraphics[width=0.88\textwidth]{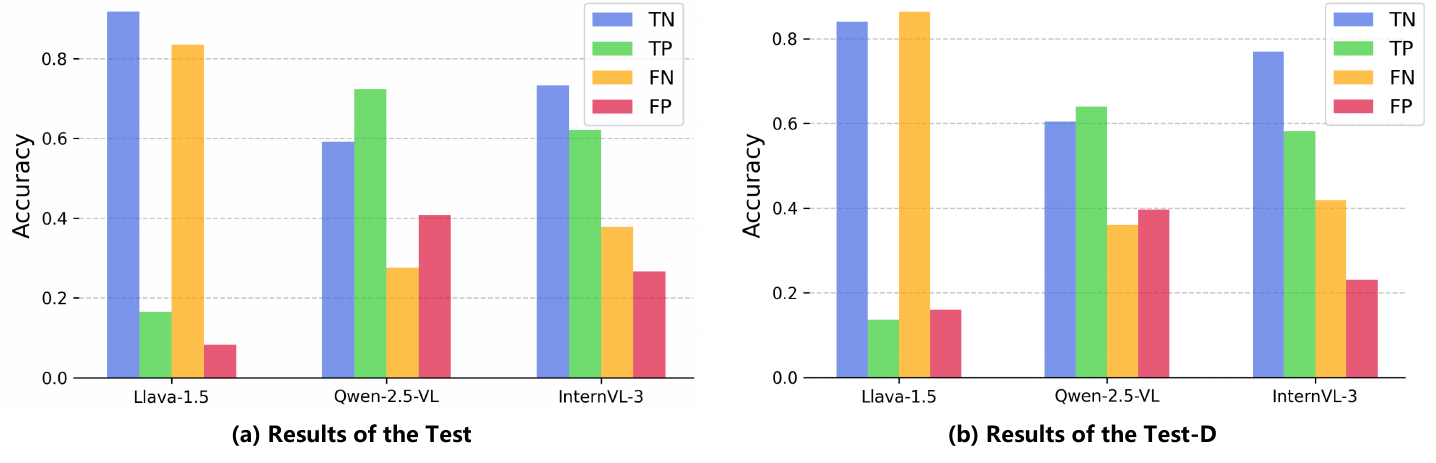}
\caption{
Zero-shot testing performance of MLLM on the MFFI Test and Test-D sets. 
True Negative (TN): Correctly identified fake samples.
True Positive (TP): Correctly identified real samples.
False Negative (FN): Real samples misclassified as fake.
False Positive (FP): Fake samples misclassified as real.
}
\label{fig:llm_1}
\end{figure*}

\subsection{Intra-dataset Evaluation}
To evaluate the overall performance of MFFI, we designed intra-dataset evaluation experiments where models were trained on the MFFI training set and tested on both the standard test set (Test) and degradation-simulated test set (Test-D). As shown in Table \ref{tab:intra-datasets_1}, on the standard Test set, the baseline Xception \cite{chollet2017xception} model achieved comparable performance to state-of-the-art detectors (RFM \cite{wang2021representative}, SRM \cite{luo2021generalizing}, SPSL \cite{liu2021spatial}), particularly attaining the highest accuracy (ACC = $0.7683$) among all methods. 
On the real-world degradation-embedded Test-D set, spatial-domain detectors (Xception \cite{chollet2017xception}, RFM \cite{wang2021representative}) demonstrated superior robustness compared to frequency-domain counterparts (SRM \cite{luo2021generalizing}, SPSL \cite{liu2021spatial}), with SRM \cite{luo2021generalizing} exhibiting the most significant ACC degradation ($\Delta$ACC = $-0.2095$). 
This indicates that the introduced degradation operations disproportionately disrupt frequency-domain features of the SRM model, highlighting the increased detection challenges of the MFFI dataset relative to existing benchmarks.

\subsection{Cross-dataset Evaluation.}
Cross-dataset testing serves as a standard evaluation protocol in face forgery detection to evaluate the generalization of the model on unseen datasets. As shown in Table \ref{tab:cross-datasets_1}, the SRM \cite{luo2021generalizing} and SPSL \cite{liu2021spatial} models exhibit suboptimal performance in cross-dataset scenarios, particularly in the highly challenging Test-D dataset. In particular, we observe that SRM \cite{luo2021generalizing} trained FF++ (C23) \cite{rossler2019faceforensics++} demonstrates superior generalizability, with trained counterparts in DF40-FS \cite{yandf40} achieving an improvement in accuracy of $0.0974$. Although DF40-FS \cite{yandf40} encompasses more diverse forgery techniques compared to FF++ (C23) \cite{rossler2019faceforensics++}, it fails to deliver expected performance gains, potentially due to insufficient coverage in diversified authentic data and varied facial scenes. These findings validate that our proposed MFFI dataset achieves superior diversity and authenticity in real-world alignment, establishing a more challenging yet practical benchmark for face forgery detection tasks.

Furthermore, we conducted cross-dataset evaluations on widely adopted unseen test sets in face forgery detection: CDF-V1~\cite{li2020celeb}, CDF-V2~\cite{li2020celeb}, DFD~\cite{rossler2019faceforensics++}, and DFDC~\cite{dolhansky2020deepfake}. Detection models were trained on the MFFI training set and tested on these unknown benchmarks. As shown in Table \ref{tab:cross-datasets_2}, the Xception \cite{chollet2017xception} model and SOTA detectors demonstrated comparable generalization performance across all datasets. This indicates that the MFFI training set effectively enhances detection models' generalization capabilities, enabling robust responses to unseen forgery challenges in real-world scenarios.

\subsection{Zero-shot Evaluation of MLLM}
With the continuous advancement of multi-modal large models, we conducted zero-shot evaluation experiments on the MFFI benchmark to assess their performance in face forgery detection. The results are visualized in Figure \ref{fig:llm_1}.
We observe that the Llava-1.5 \cite{liu2023llava} model exhibits abnormally high TN and FN values, indicating a strong bias toward classifying all samples as forged. In contrast, Qwen-2.5-VL \cite{Qwen2.5-VL} and InternVL-3 \cite{zhu2025internvl3exploringadvancedtraining} demonstrate relatively balanced performance, yet their overall accuracy (ACC) remains below 0.68. This underperformance, compared to specialized small models like SRM, fails to reflect the expected advantages of large models. These findings highlight that our MFFI dataset retains significant real-world challenges even for advanced multi-modal large models.

\section{Conclusion, Board Impact, and Limitation}

\textbf{Conclusion:}
This study introduces the Multi-dimensional Face Forgery Image (\textbf{MFFI}) dataset to address real-world face forgery detection challenges. By integrating four dimensions: Wider Forgery Methods, Varied Facial Scenes, Diversified Authentic Data, and Multi-level Degradation Operations, our MFFI effectively bridges critical gaps in scene complexity and diversity compared to existing datasets. 
The dataset encompasses $50$ distinct forgery methods and over $1024K$ image samples, demonstrating significant scale advantages. 
Experimental results show that MFFI outperforms state-of-the-art benchmarks in scene complexity, cross-domain generalization, and gradient-based detection difficulty. These findings validate MFFI’s technical innovation and practical value in advancing deepfake detection technologies.

\textbf{Broader Impact:}
The release of MFFI provides the research community with a real-world-aligned benchmark for face forgery detection while promoting the development of robust detection techniques to safeguard societal trust. The collection of raw data strictly adheres to source dataset licenses and regulatory guidelines, with usage agreements required during subsequent open-sourcing to ensure privacy protection and standardized data utilization.

\textbf{Limitation:}
Despite its large-scale and diverse coverage, the current version is limited to binary labels and excluded text annotations, which remains a critical domain. 
Future work will incorporate fine-grained forged text annotations to facilitate the development of interpretable forgery detectors based on multi-modal large models.

\begin{acks}
This work was supported by the Ant Group Postdoctoral Programme.
This work was also supported by the National Research Foundation Singapore, under its AI Singapore Programme (AISG Award No: AISG3-RP-2024-033), Fundamental Research Funds for the Central Universities (No. PA2025IISL0113, JZ2025HGTB0227), and National Natural Science Foundation of China (No. 62302139).
\end{acks}

\bibliographystyle{ACM-Reference-Format}
\bibliography{bib/diao,bib/mfl,bib/iml,bib/others,bib/sfl,bib/dfd}

\end{document}